\documentclass[pdflatex,sn-mathphys]{sn-jnl}

\jyear{2021}%

\theoremstyle{thmstyleone}%
\theoremstyle{thmstyletwo}%

\theoremstyle{thmstylethree}%

\algrenewcommand\textproc{}
\usepackage{xcolor}

\usepackage{graphicx}%
\usepackage{multirow}%
\usepackage{amsmath,amssymb,amsfonts}%
\usepackage{amsthm}%
\usepackage{mathrsfs}%
\usepackage[figuresright]{rotating}%
\usepackage[title]{appendix}%
\usepackage{xcolor}%
\usepackage{textcomp}%
\usepackage{manyfoot}%
\usepackage{booktabs}%
\usepackage{algorithm}%
\usepackage{algorithmicx}%
\usepackage{algpseudocode}%
\usepackage{program}%
\usepackage{listings}%
\usepackage{caption}
\usepackage{polyglossia}
\usepackage{float}

\begin{document}

\title[]{`MONOVAB’: An Annotated Corpus for Bangla Multi-label Emotion Detection}


\author[1]{\fnm{Sumit Kumar} \sur{Banshal}}\email{sumitbanshal06@gmail.com}

\author[2]{\fnm{Sajal} \sur{Das}}\email{sajal15-12381@diu.edu.bd}

\author[2]{\fnm{Shumaiya Akter} \sur{Shammi}}\email{shumaiya15-12179@diu.edu.bd}

\author[2]{\fnm{Narayan Ranjan} \sur{Chakraborty}}\email{narayan@daffodilvarsity.edu.bd}

\author[3]{\fnm{Vedika} \sur{Gupta}}\email{vedika.nit@gmail.com}

\author*[1]{\fnm{Mousumi} \sur{Karmakar}}\email{moukarbhu@gmail.com}

\affil*[1]{\orgdiv{Department of CSE}, \orgname{Alliance University}, \orgaddress{\street{Bengaluru}, \city{Karnataka 562106}, \country{India}}}

\affil[2]{\orgdiv{Department of CSE}, \orgname{Daffodil International University}, \orgaddress{\street{Daffodil Smart City}, \city{Birulia 1216}, \country{Bangladesh}}}

\affil[3]{\orgdiv{IS and Analytics,Jindal Global Business School}, \orgname{ OP Jindal Global University}, \orgaddress{\street{Sonipat}, \city{Haryana 131001}, \country{India}}}


\abstract{In recent years, Sentiment Analysis (SA) and Emotion Recognition (ER) have been increasingly popular in the Bangla language, which is the seventh most spoken language throughout the entire world. However, the language is structurally complicated, which makes this field arduous to extract emotions in an accurate manner. Several distinct approaches such as the extraction of positive and negative sentiments as well as multiclass emotions, have been implemented in this field of study. Nevertheless, the extraction of multiple sentiments is an almost untouched area in this language. Which involves identifying several feelings based on a single piece of text. Therefore, this study demonstrates a thorough method for constructing an annotated corpus based on scrapped data from Facebook to bridge the gaps in this subject area to overcome the challenges. To make this annotation more fruitful, the context-based approach has been used. Bidirectional Encoder Representations from Transformers (BERT), a well-known methodology of transformers, have been shown the best results of all methods implemented. Finally, a web application has been developed to demonstrate the performance of the pre-trained top-performer model (BERT) for multi-label ER in Bangla.}




\maketitle

\section{Introduction}\label{sec1}

Sentiment Analysis (SA), more precisely aspect-based sentiment analysis, is a branch of Natural Language Processing (NLP) that evaluates whether a given text represents any sentiment or not\footnote{\url{https://monkeylearn.com/sentiment-analysis/}}. The types of sentiments were relatively paid more attention to analyze positive, negative or neutral sentiments. ER is another processing type of NLP that aids in determining the specific emotion or emotions represented. Identifying particular emotions can help to visualize the broader sense of a given text where SA may provide the vibe of that feelings \cite{Tripto2018}. For example, a text may have a positive connotation, but this does not provide an explicit idea about the emotion of the text. Positive sentiments include some particular emotions such as being happy, joyful, hopeful, etc. ER helps to comprehend the specific spirit of a sentence. Thus, it has a broader implication to understand the outcome of a given text in any preferred aspect \footnote{\url{https://recfaces.com/articles/emotion-recognition}}.

Several studies have been conducted for the purpose of recognizing emotions in the Bangla language \cite{Tripto2018,Rahman2020,Khan}. The majority of studies focus on identifying some particular emotions \cite{144,Purba,Pal2020}. Anger, contempt, disgust, enjoyment, fear, sadness, and surprise are the commonly identified emotions \cite{Khan}. However, there are always several scenarios where a single text can convey a variety of feelings \cite{Tripto2018}. For example, anger and disgust can be expressed in the same utterance and this task can be employed by multi-label ER. Because multi-label ER helps to extract several emotions from a single text \cite{140}. But, relatively very less focus has been paid on this concern area. Thus, multi-label emotion detection might be an interesting approach in order to mitigate the challenges of this kind.

There is a deficiency of the development in the area of multi-label emotion recognition in the Bangla language \cite{Tripto2018,Rayhan}. Because the availability of proficiently labeled data as well as suitable tools and approaches are penurious in this area \cite{Kowsher2021,Tasnim2021}. For that reason, Bangla is still debated as a language with limited resources or low resource language \cite{144}. A comprehensive literature study has been conducted where the limitations of this domain have been outlined and a generalized framework has been extracted to acknowledge the existing tools and techniques \cite{144}. The following criteria have been considered for inclusion in this research as an extension of the previous literature study:

\begin{enumerate}[1.]
\item How can a context-based emotion corpus be compiled for Bangla?

\item How multi-label emotions can be pre-processed and identified from context-based annotated corpus for Bangla?
\end{enumerate}
This research work focuses on the most crucial aspect of emotion detection of Bangla text, i.e., a well-annotated compiled annotated corpus. To do so, an automated tool has been developed for collecting the data. Based on the context of the data, A number of annotators have contributed (native speakers) to compiling the same. Furthermore, the annotated data has been processed through the generalized framework of processing Bangla text as per established \cite{144}. Using the findings of multilabel emotions and the annotated corpus, a user interface-based web application has been developed. 

\section{Related Work}\label{sec2}
The usage of SA and ER has seen substantial growth due to their broader implications in several domains. With the penetration of online media, attracts eyeballs irrespective of the domain. With the same propagation, researchers have also been attracted towards SA or ER-based works over the period \cite{history1}. Among the earliest works, “The Cross-Out Technique as a Method in Public Opinion Analysis” was introduced in the 1940s \cite {history2}. Adaption of computers in different spheres also broadens the implication of the research field around the 1990s \cite{137}. However, phenomenal research growth has been observed from 2004 onwards. Emotions tend to reflect through expressions, and gestures rather than only from speech or text. Thus image processing for detecting emotions implementing deep learning techniques was found to more profound aspect of ER \cite{141}. Visual representation of gestures thus stimulates through the image processing field with the availability of tools \cite{141,142,Alam2020,143}. The growth of abundant text data especially from social media has shaped the domain into an attractive research area afterwards \cite{history1, rev92022, rev102022, rev112022, history3}.

However, Jargons and written form of Bangla text widely differ from place to place that requires different NLP task to handle each data. These variations thus imply strategizing the process of recognizing the characteristics of underlying text \cite{rev12022, rev22022, rev32022}. But the explosive growth of social media enables people across borders to communicate their views, thoughts, discussions moreover emotions in the form of text. In this era of online evolution, companies grab these as an opportunity to understand consumer preferences \cite{Bhargava2019, Ahmad2019}. The market-oriented implications of social media emotion recognition make it one of the most trendiest topics in NLP tasks \cite{Pal2020,Bhargava2019,Miranda2018}. Among all the languages, English has an abundant presence in social media \cite{Dawn,Mahmud2019} being considered as an international language \cite{Noor2019} most likely widens this presence. Several tools and techniques for this language are also available \cite{Mandal2018,Chathumali,Hossain2019a}. Because various mechanisms are basically built for this language, and this is why comparatively easier to train machines using data from the English language \cite{Ghosh2017,Hawladar2021,Das2020,Batbaatar2019}. Apart from the ease of training, the abundance of pre-processing tools makes this domain more accessible and explored \cite{Islam2017,Mahmud2019,Hawladar2021} \cite{144}. Consequently, the empirical evidence of the efficacy of algorithms like Support Vector Machine (SVM), Naive Bayes (NB), Multinomial Naive Bayes (MNB), Logistic Regression (LR), K Nearest Neighbor (KNN), and Random Forest (RF) noted more profoundly \cite{144, Sarowar2019,Sharif2019,Noor2019, Ullah2020}. Similarly, algorithms such as Long-Short Term Memory (LSTM), Bidirectional Long-Short Term Memory (BiLSTM), CNN, and some hybrid algorithms are popularly used \cite{144, Sarkar2021,Ullah2020,Deshpande2018, HOSSAIN20211}. Along with traditional approaches rule-based, transformer-based, lexicon-based, stacked ensemble, and word embedding methods were recorded as useful for this language in different datasets \cite{144, Sazzed2020,Sarkar2019, Rabeya2018, Gupta2021}.

In contrast, Bangla despite being the seventh most spoken language of the world, is arguably still considered a resource-constrained language in terms of analytical landscape. One prominent reason behind that might be due to the structural and dialectical complexity of the same \cite{Rahman2020,Khan2020, 144}. Despite all complexities, Bangla is widely adopted in different social media (e.g. Facebook) and online (etc. e.g. Google) platforms due to its popularity among people around the world. Thus, the generation of the language in textual form has seen extraordinary growth over time \cite{Khan2020,Pal2020}. Eventually, the availability of such data propagates the domain of SA and ER imposed in the Bangla language as well. The notable works started from determining positive or negative disposition \cite{Dey2019} to identify a particular \cite{Basri2021,Tuhin2019} or multiple emotions \cite{Tripto2018}. Moreover, numerous works have been conducted on the subject of multi-class emotion recognition \cite{Asimuzzaman2018, 144}. In this process of multi-class ER, the most popular approaches for Bangla were found to be Rule-based, DL-based or Word Embedding based works \cite{Banik2018,Hoque2019,RafidulHasanKhan2021}.

Along with identifying the multi-class emotions, to start with the process, the pre-processing techniques also needed to be identified. Typically, tokenization was noted to be used as one of the most used techniques for processing the language \cite{144}. For extracting features usual techniques like n-gram, TF-IDF etc were implemented \cite{144, Asimuzzaman2018, Azmin2019}. Though, the deficiency of appropriate tools in processing the same is always been a concern in handling Bangla language texts \cite{Saha2020, Haque2020a}. Nevertheless, a transformer-based method for Bangla known as BERT developed and gained popularity in the recent past \cite{Islam2020, Das2021, Song2021, AbhikB}. In the case of identifying multiple emotions, Ekman’s emotions were profoundly used in previous works due to their relevance and inclusiveness of probable emotions in the Bangla language \cite{Khan,Basri2021,144,Tuhin2019}.

Multi-label ER, which is an acronym that stands for the detection of multiple emotions from a given text, recently gained popularity among researchers \cite{Tripto2018,Rayhan} in order to classify human feelings more precisely. Nonetheless, recognizing multiple feelings conveyed is an arduous task \cite{Tripto2018}. Due to this fact, relatively fewer attempts have been made to address this issue in other languages as well. So, to tackle this problem, this approach of multi-label emotion identification has been developed from social media platforms. In this process, one by-product is produced as the collection procedure of the data from social media platforms. Finally, a web app has been implemented to use the verified methodology for a given text. Where the tool can identify the emotions conveyed with precision based on contextual understanding.

\section{Experimental Framework}\label{sec3}
To alleviate the challenge of identifying emotions from the Bangla language, a broader experimental setup has been designed and adopted in this study. The experimental framework has three main segments: scraping and preparing the data, identifying aspect-based multi-label emotions and finally a web application.

\subsection{Dataset Preparation}\label{subsec3.1}

In the process of evaluating any algorithm, the relevance, abundance, structure and volume of data play a crucial role. Typically, for recognizing emotions, social media, reviews, newspapers etc are preferred platforms due to diversity and inclusiveness \cite{144, Sazzed, Gupta2}. Bangladesh is the most prominent region consisting of Bangla-speaking people and Facebook is the most leading social media platform in this region \cite{Mandal2018}. Also, it is a great source of text data used for various purposes including emotion extraction \cite{Sazzed2020}. Due to its popularity, it includes views from wider society to broaden the understanding of a given context. Additionally, the news portal provides context which might draw different emotions across a society  \cite{Das2021, Sarkar2021}. Facebook is also a very viable source in this aspect as it includes a wider range of news portals in Bangla. Based on these ideas of heterogeneity and comprehensiveness, six popular news portals \cite{Rahman2020, Sharmin2020} have been selected. These news portals are present on Facebook and recorded in some earlier works as well \cite{Hossain2017, Sharmin2020}. To collect data from these news portals, Eleven News spreads widely in Bangladesh during 2018-2022 on social and religious issues. 
Eleven News among the top discussed was on the following contexts: 
\begin{enumerate}[1.]
\item Conflict between College Students and Businessmen\footnote{\url{https://www.facebook.com/DailyProthomAlo/posts/5792302497469455}}.
\item An Accused School Teacher of Religious Contempt\footnote{\url{https://www.facebook.com/BBCBengaliService/posts/347783877392906}}. 
\item Student Activism in Demand of Safe Road Transport\footnote{\url{https://www.facebook.com/DailyProthomAlo/posts/2156727967693611}}. 
\item Student Protest on Quota Reformation in Job Sector\footnote{\url{https://www.facebook.com/DainikJugantor/posts/2043025162389103}}. 
\item Unusual Price Hike of Refined Soyabean Oil\footnote{\url{https://www.facebook.com/DailyProthomAlo/posts/5642953425737697}}. 
\item Sudden Strike of Railway Employees and Stranded Passengers\footnote{\url{https://www.facebook.com/bdnews24/posts/10166576508225601}}. 
\item University Student Protest in Demand of Vice-Chancellor of University\footnote{\url{https://www.facebook.com/DailyProthomAlo/posts/5513602728672768}}. 
\item Minority Oppression During Their Biggest Religious Festival "Durga Puja"\footnote{\url{https://www.facebook.com/BBCBengaliService/posts/4689559891082879}}. 
\item Religiousization of ``Mangal Shobhayatra" on Bengali New Year\footnote{\url{https://www.facebook.com/DailyProthomAlo/posts/5771832749516430}}.
\item Harassment of a Female School Teacher for Wearing Bindi\footnote{\url{https://www.facebook.com/DailyProthomAlo/posts/5747031511996554}}. 
\item Suspension of an Employee from Anti-Corruption Commission (ACC)\footnote{\url{https://www.facebook.com/dailymanabzamin/posts/4891037690982429}}. 
\end{enumerate}

To prepare the dataset, a scraping algorithm has been developed to collect the data about mentioned news from Facebook. The comments made by users on Facebook pages on these News were collected. After initial pre-processing, it is annotated to build a labeled corpus for comprehensive emotion recognition.

\subsubsection{Scraping Algorithm}
\begin{figure}[h]%
\centering
\includegraphics[width=0.9\textwidth]{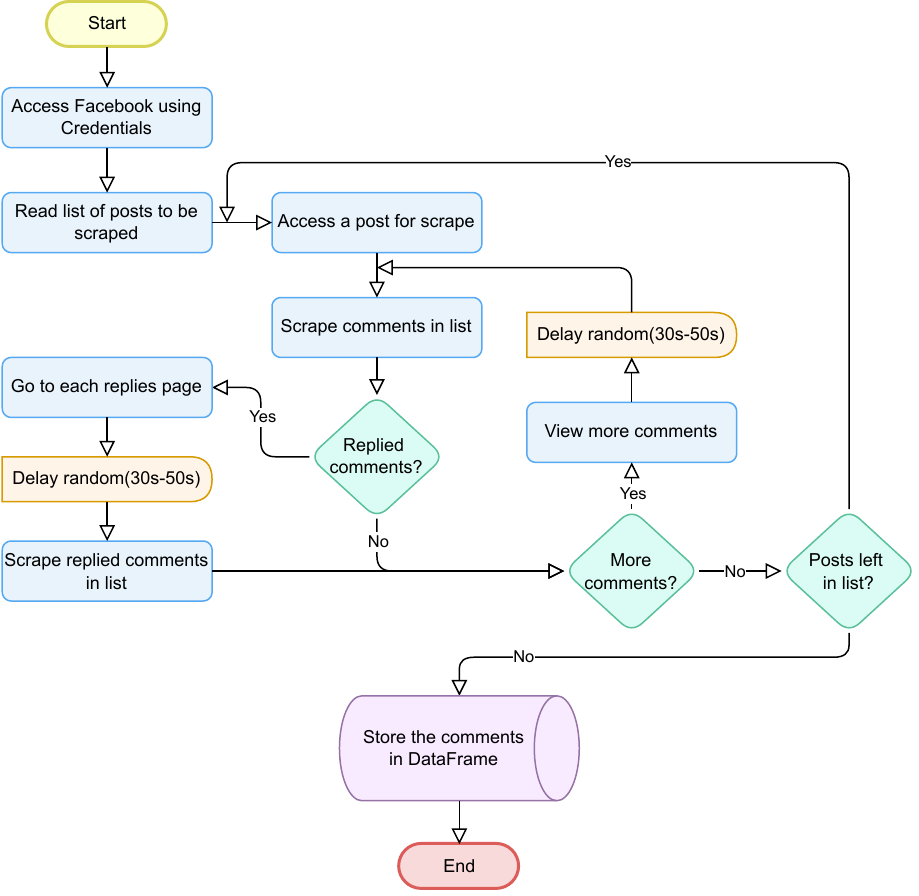}
\caption{Flowchart of scraping algorithm}\label{fig1}
\end{figure}

To automate the process of data collection, a scraping algorithm has been implemented focusing on Facebook pages. The workflow diagram of the same has been depicted in \autoref{fig1}. The scraper tool was designed to start with the link to the Facebook post of given news. It fetched any comments made on the post and further replies made on the comment. In this process of fetching a random delay of 30-50s was added not to overload the server. During this fetching operation, a couple of processing also implemented to remove any duplication or blank comments. All the links have been read from a file and look for respective comments. The process iterates over every comment first to fetch any replies that existed. Finally, the scraper iterated over each link provided in the file. The source code and documentation of the scraping algorithm can be accessed from GitHub \footnote{\url{https://github.com/sajaldoes/FacebookScraper}}. The number of comments for each news found is incorporated in \autoref{tab1}.

It can be observed from \autoref{tab1} that the number of comments is somehow different from one another news. The reason behind that might be the duration of the event. For example, News 6 focused Strike of Railway employees being dissolved within a few hours. Though, it created a spark within a very short time but doesn't last long. Whereas News 8 focusing the attack on the `Durga Puja Pandals' that happened over a week thus, it generates more comments. Though the social media event is about instant reaction, it still needs some time. Posts that have received less than 400 responses are removed manually from the selected list. Due to the fact that News 6 is touched by fewer people, it has the fewest number of comments. On the other hand, news 10 has the most comments since it has a wide-reaching impact on people's minds and thus generates the most discussion. Finally, a total of 136,583 pieces of comments have been extricated for the further extraction of emotions.

\subsubsection{Data Preprocessing}
Several steps of pre-processing have been performed in the transition from the raw data to well build a corpus of sentences. As the raw data has been collected from social media, thus the prominence of emoticons, symbols and pictographs existed. The focus is to work with Bangla text, thus these have been removed to clean the data. Following by some manual interventions such as removing null comments, resetting comment indexes, screening and removing irrelevant comments etc. In the final leg of pre-processing the data, handle the English characters in Bangla sentences. The most occurred characters are punctuation followed by numerals. For instance, the number 500 was converted to the Bangla digit {\bangla ৫০০}, and the digit 00 was changed to {\bangla ০০}. Finally, the English punctuation such as \texttt{\textbar } was changed to Bangla punctuation {\bangla ।}, which is used as a full stop in Bangla sentences. All this preprocessing finally resulted in a set of 1,08,950 comments to prepare an annotated corpus.

\begin{table}[h]
\begin{center}
\begin{minipage}{320pt}
\caption{News wise posts and comments ratio}\label{tab1}%
\begin{tabular}{lrrrr}
\toprule
No. & News Headline & Year  & Num. of Posts & Num. of comments \\
\midrule
1 & {\bangla নিউমার্কেট, ঢাকা কলেজ সংঘর্ষ}    & 2022     & 14                       & 11294                       \\
2 & {\bangla ধর্ম অবমাননায় কারাবন্দী শিক্ষক}    & 2022     & 13                       & 10098                       \\
3 & {\bangla নিরাপদ সড়ক আন্দোলন}    & 2018    & 13                       & 12192                       \\
4 & {\bangla কোটা সংস্কার আন্দোলন}     & 2018    & 14                       & 16582                       \\
5 & {\bangla সয়াবিন তেলের মূল্য ঊর্ধ্বগতি}  & 2022       & 15                       & 14973                       \\
6 & {\bangla রেল ধর্মঘট}    & 2022     & 3                        & 536                         \\
7 & {\bangla শাবিপ্রবি ভিসি আন্দোলন}    & 2022     & 13                       & 7302                        \\
8 & {\bangla দূর্গা পূজা সংঘর্ষ}    & 2021     & 15                       & 22930                       \\
9 & {\bangla মঙ্গল শোভাযাত্রা বিতর্ক}    & 2022     & 9                        & 7302                        \\
10 & {\bangla টিপ্ বিতর্ক}   & 2022     & 17                       & 33188                       \\
11 & {\bangla চাকরিচ্যুত দুদক কর্মকর্তা}   & 2022    & 5                        & 1765                        \\
\botrule
\textbf{Total} & & \textbf{130}             & \textbf{136583}        \\
\botrule
\end{tabular}
\footnotetext{“Year” represents the occurrence of the incident, “Num. of posts” shows the amount of posts selected for each news and “Num. of Comments” shows the total number number of comments collected for each news. For example, 14 posts are selected for first news and there are 11294 comments are collected from 14 posts.}
\end{minipage}
\end{center}
\end{table}

\subsubsection{Data Annotation}
Data annotation is the process of identifying properties of training data (such as text, images, audio, and video) to give machines information about the data's content and significance. An innovative aspect of the data annotation in this study is context-based annotation \footnote{\url{https://www.defined.ai/blog/machine-learning-essentials-what-is-data-annotation/}}. Context-based annotation refers to the annotation of data depending on the underlying context \cite{144}. Contexts were supplied to the annotators of the associated news articles during the annotation phase so that they could comprehend the emotions of comments more precisely before annotating them. Since a huge amount of data was collected for this study, the work of annotating the data required the participation of a large number of individuals. Accordingly, the key focus of this research is the accuracy of the annotation. Due to this, three annotators were tasked with labeling a single batch (200 comments) of data to ensure the data was tagged with the proper understanding of context. This article identifies five emotions, including anger, contempt, disgust, enjoyment, and sadness. Since the genesis of the material is some severe, widespread concerns, these feelings are appropriate for the acknowledgment. \autoref{tab2} presents the annotation of some data for multi-label emotional states.

\begin{table}[h]
\caption{Multi-label emotion annotation sample data}\label{tab2}%
\begin{tabular}{p{4.2cm} p{2cm} p{2cm} p{2cm}}
\toprule
Comment  & Annotator 1 & Annotator 2 & Annotator 3 \\
\midrule
{\bangla শোষিত জনগন এইটাই আশা করেছিল!!!}         & Anger, Disgust                       & Anger, Disgust & Disgust, Sadness                       \\
{\bangla নেও ঈদের আগে দোকান বন্ধ করে বসে থাক।}         & Anger, Contempt                      & Contempt, Disgust & Contempt, Disgust                      \\
{\bangla দেখার কি কেউ নেই।}        & Sadness                       & Sadness & Contempt                     \\
{\bangla অসাধারণ! প্রসংশা করার ভাষা হারিয়ে ফেলেছি।}
         & Enjoyment                      & Enjoyment   & Enjoyment                   \\
{\bangla এদের পাগলের সাথে তুলনা করলে পাগলরা অপমানিত হবে।}       & Anger, Contempt                       & Anger & Anger, Disgust                      \\
\botrule
\end{tabular}
\footnotetext{Annotator 1, Annotator 2, Annotator 3 refers to the labeled emotions by each annotator where annotator 1, 2 labeled the first comment as anger and disgust and annotator 3 denoted it as disgust and sadness emotions.}
\end{table}

\begin{figure}[h]%
\centering
\includegraphics[width=0.9\textwidth]{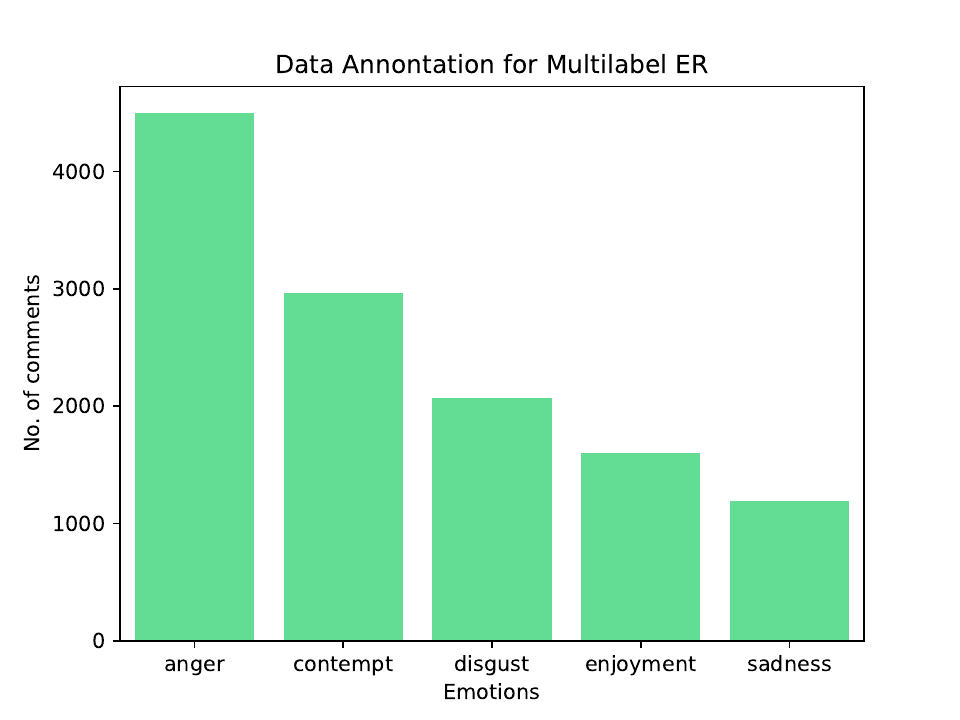}
\caption{Data Annotation for multi-label emotion recognition}\label{fig3}
\end{figure}

After finishing the step of data annotation, this article has conducted some preprocessing on the acquired data to refine it. In this phase, an alternative regular expression was used to delete any identified superfluous or unneeded punctuation. In addition, longer than two-word comments were included during model evaluation.

\subsubsection{Feature Extractions}
For the extraction of features, this study used tokenization and Term Frequency Inverse Document Frequency (TF-IDF) approaches. Tokenizers were applied for text-to-number conversion  \cite{Hossain2020}. Since computers cannot interpret human language, the data must be transformed into a format that computers can understand. Several research \cite{Banik2018,Amin2019} employ tokenization as a preprocessing strategy for emotion identification. However, DL employs tokenizers as a feature extraction approach to turn the data into features before feeding them as input to the model  \cite{Saha2021,Asani2021}. Consequently, tokenization was used in this study to perform feature extraction for DL methods. The function of the tokenizer was determined by two parameters, including the number of words to retain based on their frequency and tokens for non-vocabulary terms. The tokenizer was then used to text to generate sequences from the data. In addition, these sequences are transformed into pad sequences in order to give a full feature for the implementation of the DL model.

TF-IDF, on the other hand, examines the importance of a word inside a phrase or text based on how frequently and infrequently it occurs and departs \cite{Khan}. TF-IDF is a statistical metric that assesses the significance of a phrase inside a corpus document. It is based on the accumulation of a phrase's frequency in a document but is balanced by the number of papers that contain the term. This is accomplished by multiplying two metrics: the frequency with which a word appears in a phrase or document and the inverse document frequency, which is the frequency with which the term appears throughout a corpus of phrases and documents \cite{Sarowar2019}. Throughout this research work, the TF-IDF vectorizer plays a crucial role in the development of machine-learning algorithms. A vocabulary is then generated using the TF-IDF vectorizer, which uses a tri-gram range. The vectorizer then utilized the 5000 words to generate the characteristics. The minimum and maximum limits of the tri-gram range are (1,3). Moreover, the vectorizer transforms the input and generates a normalized TF-IDF feature matrix for the model. In this work, therefore, these strategies are used for feature extraction.

\subsection{Experimental Analysis}\label{subsec3.3}

Various conventional methods have been analyzed and implemented in order to identify emotions in several research \cite{Wahid2019,Shafin}. This study employed ML, DL, and transformer-based algorithms for the detection of multiple types of emotional expressions. The transformer-based technique is one of the rule-based approaches that outperforms all other algorithms \cite{Das2021}. 
Using context-based data, the succeeding subsections will present examples of the applicable approaches in various application domains.

\subsubsection{Conventional  Approaches}\label{subsubsec3.3.1}
LR, RF, MNB, SVM, and KNN have been evaluated from the conventional ML domain which is frequently implemented to identify emotions \cite{Purba,Ghosh2020, 144}. The corresponding outcomes of several ML approaches are demonstrated in \autoref{tab3}.

Logistic Regression (LR):
LR is a category of supervised ML models \cite{Hoque2019,Sazzed,Hossain2017, DasDawn2020}. LR assesses the likelihood that an occurrence, such as voting or not voting, will happen based on a specific number of predictor variables \cite{MontaserHamid2018,Saha}. 

\begin{equation}
p(x)=\frac{1}{(1+e^{-(\beta_0+\beta_1 x})}
\end{equation}

\begin{sidewaystable}
\sidewaystablefn%
\begin{center}
\begin{minipage}{\textheight}
\caption{Precision, recall, and F1 score of ML approaches}
\label{tab3}
\setlength{\tabcolsep}{4.5pt} 
\renewcommand{\arraystretch}{1} 
\begin{tabular*}{\textheight}{lccccc|ccccc|ccccc}
\toprule%
& \multicolumn{5}{c}{Precision (\%)}& \multicolumn{5}{c}{Recall(\%)} & \multicolumn{5}{c}{F1 Score} \\\cmidrule{2-6}\cmidrule{7-11}\cmidrule{12-16}%
Emotions & LR & MNB & SVM & RF & KNN &	LR & MNB & SVM & RF & KNN & LR & MNB & SVM & RF & KNN \\
\midrule
\textbf{Anger}                                         & \textbf{70.59} & \textbf{74.84} & 73.02        & 71.69       & 64.16        & \textbf{70.01} & 66.63        & 65.69          & 64.41          & 43.87        & \textbf{0.7}  & \textbf{0.7}  & 0.69          & 0.68        & 0.52         \\
\textbf{Contempt}                                      & 60.47          & \textbf{72.84} & 66.08        & 63.74       & 41.48        & 52.77          & 42.07        & \textbf{54.98} & 49.63          & 45.39        & 0.56          & 0.53          & \textbf{0.6}  & 0.56        & 0.43         \\
\textbf{Disgust}                                       & 40.49          & \textbf{64.29} & 51.96        & 50.0          & 25.54        & \textbf{28.4}  & 8.89         & 13.09          & 12.1           & 17.53        & \textbf{0.33} & 0.16          & 0.21          & 0.19        & 0.21         \\
\textbf{Enjoyment}                                     & 68.85          & \textbf{89.94} & 64.09        & 62.89       & 27.7         & 70.2           & 59.61        & \textbf{74.9}  & 71.76          & 57.25        & 0.7           & \textbf{0.72} & 0.69          & 0.67        & 0.37         \\
\textbf{Sadness}                                       & 36.16          & \textbf{90.91} & 32.0         & 23.64       & 14.63        & 30.62          & 4.78         & 34.45          & \textbf{35.41} & 8.61         & \textbf{0.33} & 0.09          & \textbf{0.33} & 0.28        & 0.11   \\
\botrule
\end{tabular*}
\footnotetext{Bold values indicate the highest precision, recall, and F1 score among the algorithms for the classified emotions.}
\end{minipage}
\end{center}
\end{sidewaystable}

In this model, the beta parameter or coefficient is typically determined via maximum likelihood estimation (MLE). \footnote{\url{https://www.ibm.com/topics/logistic-regression}}. Furthermore, the stopping criterion tolerance is set to 1e-4, and the inverse regularization strength is set to 12.0 units. This model produced the greatest precision score for anger emotion, the highest recall for the enjoyment emotion, and the highest f1 score of 0.7 for both anger and enjoyment.

Random Forest (RF):
RF is a strong and flexible supervised ML technique that develops and aggregates several decision trees to generate a ``forest" \cite{Pal2020}. In this study, RF is used to identify emotions by employing 10 trees as estimators and the Gini impurity criterion. The maximum number of data features is obtained by calculating the square root of the number of data features. Additionally, RF constructs the forest via its base estimator decision tree classifier strategy. This algorithm's accuracy in predicting anger, contempt, and delight emotions approached 60 percent, but disgust stayed at 50\% and sorrow dipped below 30 percent. In contrast, the delight class has the highest remember value while the disgust class has the lowest recall value. However, the total f1 score reveals that anger, contempt, and delight emotions were identified better than the others.

Multinomial Naive Bayes (MNB):
The MNB classifier is a statistical approach that is utilized in NLP most of the time \cite{Hasan2019,Mahmud2019,Purba}. The probabilities of this model were developed by Thomas Bayes, and it can be stated as follows:

\begin{equation}
    P(\frac{A}{B})=\frac{P(A \cap B)}{P(B)} =  \frac{P(A)*P(\frac{B}{A})}{P(B)}
\end{equation}

 As per the empirical observance, MNB is one of the best-performing ML models. This algorithm’s multinomial distribution accurately categorizes all five emotions with a precision score of more than 60\%. Sadness emotion has the greatest precision score of 90.91\%, while disgust has received the lowest precision score of 64.29\%. Furthermore, it has performed in terms of recall across all emotions other than disgust and sadness.

Support Vector Machine (SVM):
A linear model that can be used for classification and regression issues is known as SVM \footnote{\url{https://towardsdatascience.com/https-medium-com-pupalerushikesh-svm-f4b42800e989}}. To train the data, the 1e-3 tolerance for halting criterion rate is used, and the one versus rest decision function is used to predict emotions from the text. High levels of precision scores have been observed for all emotions except sadness. In terms of recall, disgust has been very less identified whereas, the emotion of enjoyment has been identified more proficiently. Similar patterns have been observed in terms of f1 score among anger, contempt, and enjoyment emotions.

K Nearest Neighbor (KNN):
The k-nearest neighbor method, generally known as KNN, is a non-parametric, supervised learning technique that leverages proximity to make predictions or classifications regarding the grouping of a single data point \cite{Ghosal2016,Khan,Mahmud2019}. In this study, this system performed poorly in categorizing emotions with multiple labels. Only in the instance of anger, the precision score recorded above 50 percent, although for enjoyment, the recall score was 57.25 percent, and both anger and enjoyment showed promising identification on the scale of f1 score for this model.

\begin{figure}[h]%
\centering
\includegraphics[width=0.9\textwidth]{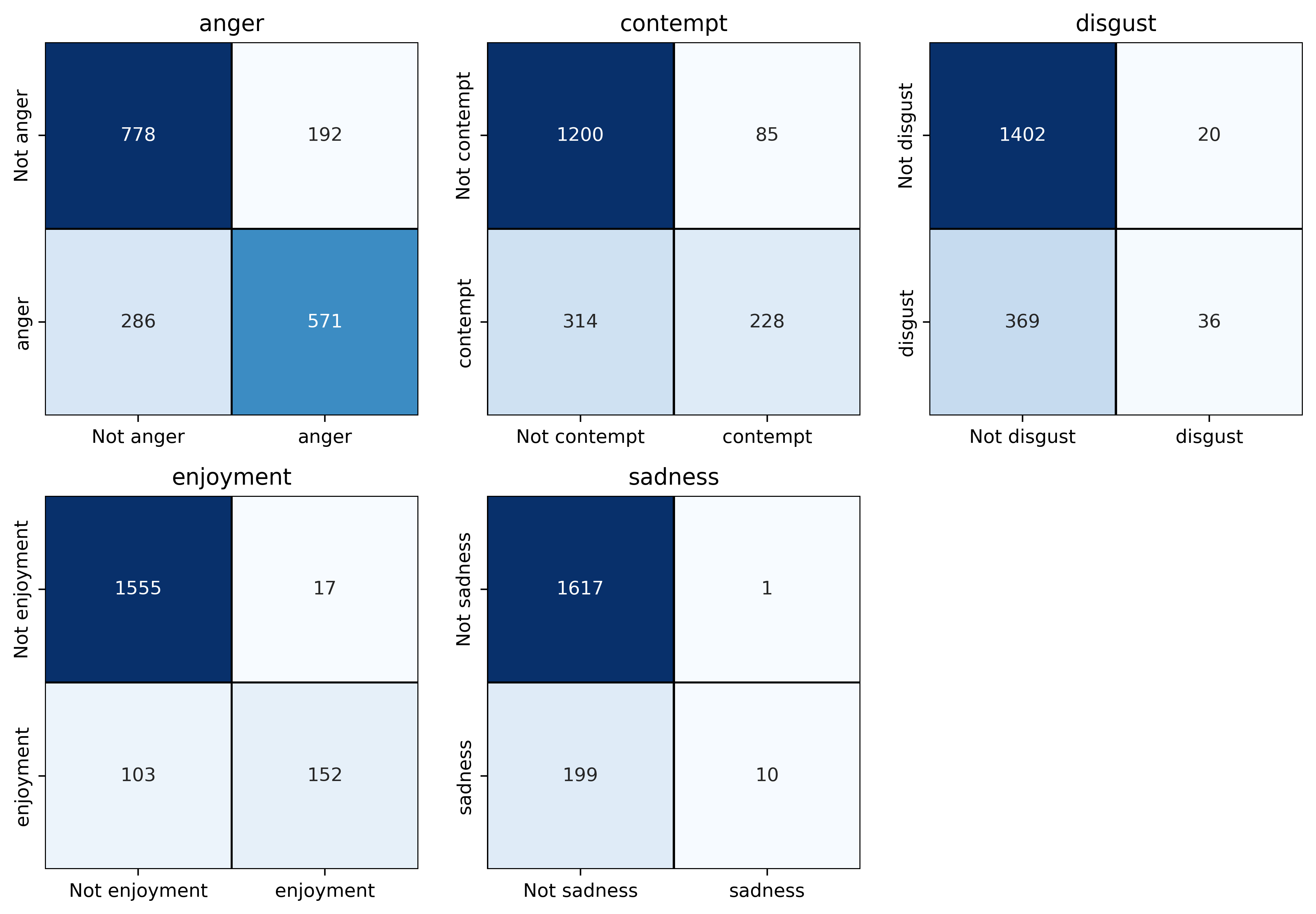}
\caption{Confusion matrix of MNB that is best performer among ML approaches. Each box indicates the number of correctly and incorrectly identified emotions.}\label{fig4}
\end{figure}

\begin{table}[h]
\begin{center}
\begin{minipage}{100pt}
\caption{Overall accuracy score of ML algorithms}\label{tab6}
\begin{tabular}{lr}
\toprule
Model & Accuracy (\%) \\
\midrule
LR             & 80.03                  \\
\textbf{MNB}   & \textbf{82.64}         \\
SVM            & 80.72                  \\
RF             & 79.11                  \\
KNN            & 70.87                 \\
\botrule
\end{tabular}
\footnotetext{Bold value specifies the highest result among ML algorithms.}
\end{minipage}
\end{center}
\end{table}

Consequently, these ML models are incorporated into the classifier chain approach, which improves the effectiveness of classifying multi-label emotions. \autoref{tab6} is a representation of the computational outcomes found while employing ML algorithms. The MNB algorithm achieves the greatest performance among all ML methods. The documented accuracy of this procedure is 82.64 percent.  \autoref{fig3} displays the algorithm's confusion matrix for improved comprehension and visualization. Although this strategy produces higher outcomes, it does not give the maximum performance across all categories. The top performance in this challenging undertaking will be addressed in further depth in the next section.

\subsubsection{Evaluation of DL approaches}\label{subsubsec3.3.2}
DL algorithms are frequently employed in the categorization of text \cite{Leonova2020}. Despite the fact that this domain is ideally suited for image processing, it is gaining popularity for text-based ER in a variety of languages \cite{Hossain2020a,Rabby2021,Sarkar2021}. Some common algorithms such as LSTM, BiLSTM, CNN-LSTM, and CNN-BiLSTM are utilized in this work \cite{144}. Different scoring metrics for each method, including accuracy, recall, and F1 score, are displayed in \autoref{tab4}..

Long-Short Term Memory (LSTM):
LSTM networks are designed primarily to solve the issue of long-term dependability \cite{Sazzed}. It does not merely apply the past forecast, but rather preserves a long-term context in memory, which helps it avoid the long-term dependency problem that other models experience \cite{Salehin2020}. A four-layer LSTM model is applied in order to identify the emotions communicated by this work. Combining the embedding layer with the input length set to 100, embedding size set to 300 units, and vocabulary size chosen by the tokenizer based on the data produces a sequential model. After that, the model's major layers are formed using two LSTM layers containing 128 and 64 neurons, and it is completed with an output-dense layer containing a single neuron and sigmoid activation function. In addition, the model was trained using an adam learning rate that is three times 1e-4. This LSTM model is employed for each label in order to train it particularly as a classifier chain technique, resulting in an increase in the accuracy of multi-label classification.

Bidirectional Long-Short Term Memory (BiLSTM):
BiLSTM is most often employed for NLP \cite{Hoque2019,Hossain2020a}. In contrast to standard LSTM, this one allows input from both ways and may utilize data from both sides \cite{Rayhan}. Four-layer BiLSTM model is used to determine the emotions represented by this work. Combining the embedding layer with the input length set to 100, embedding size set to 300 units, and vocabulary size chosen by the tokenizer based on the data creates a sequential model. Then, two bidirectional LSTM layers with 128 and 64 neurons and an output dense layer with a single neuron and a sigmoid activation function are used to form the major layers. This model has also been applied as a classifier chain approach, resulting in improved multi-label classification accuracy. Moreover, compared to all others, BiLSTM had the greatest recall score for the emotion of delight. In addition, it has preserved an excellent memory for each feeling.

\begin{sidewaystable}
\sidewaystablefn%
\begin{center}
\begin{minipage}{\textheight}
\caption{Precision, recall, and F1 score of DL approaches}
\label{tab4}
\setlength{\tabcolsep}{4pt} 
\renewcommand{\arraystretch}{1} 
\begin{tabular*}{\textheight}{lcccc|cccc|cccc}
\toprule%
& \multicolumn{4}{c}{Precision (\%)}& \multicolumn{4}{c}{Recall (\%)} & \multicolumn{4}{c}{F1 Score} \\\cmidrule{2-5}\cmidrule{6-9}\cmidrule{10-13}%
Emotions & LSTM & BiLSTM & \begin{tabular}[c]{@{}c@{}}CNN-\\ BiLSTM\end{tabular} & \begin{tabular}[c]{@{}c@{}}CNN-\\ LSTM\end{tabular} & LSTM & BiLSTM & \begin{tabular}[c]{@{}c@{}}CNN-\\ BiLSTM\end{tabular} & \begin{tabular}[c]{@{}c@{}}CNN-\\ LSTM\end{tabular} & LSTM & BiLSTM & \begin{tabular}[c]{@{}c@{}}CNN-\\ BiLSTM\end{tabular} & \begin{tabular}[c]{@{}c@{}}CNN-\\ LSTM\end{tabular} \\
\midrule
\textbf{Anger}                     & \textbf{70.72} & 68.88           & 67.25                                                          & 67.81             & 55.77          & 53.37           & \textbf{61.03}                                                 & 58.74             & 0.62          & 0.6             & \textbf{0.64}                                                  & 0.63              \\
\textbf{Contempt}                  & 61.74          & \textbf{65.54}  & 61.95                                                          & 61.33             & 41.94          & 41.61           & \textbf{48.19}                                                 & 44.08             & 0.5           & 0.51            & \textbf{0.54}                                                  & 0.51              \\
\textbf{Disgust}                   & \textbf{34.8}  & 33.88           & 30.85                                                          & 30.85             & \textbf{42.03} & 30.19           & 35.02                                                          & 28.99             & \textbf{0.38} & 0.32            & 0.33                                                           & 0.3               \\
\textbf{Enjoyment}                 & 65.04          & 60.0            & \textbf{66.98}                                                 & 52.28             & 71.61          & 74.76           & 67.82                                                          & \textbf{76.03}    & \textbf{0.68} & 0.67            & 0.67                                                           & 0.62              \\
\textbf{Sadness}                   & 33.68          & \textbf{41.61}  & 31.25                                                          & 40.74             & \textbf{42.24} & 24.57           & 17.24                                                          & 14.22             & \textbf{0.37} & 0.31            & 0.22                                                           & 0.21                \\
\botrule
\end{tabular*}
\footnotetext{Bold values indicate the highest precision, recall, and F1 score among the algorithms for the classified emotions.}
\end{minipage}
\end{center}
\end{sidewaystable}

Hybrid Deep Learning Approaches: 
Four separate layers are used to generate CNN-BiLSTM and CNN-LSTM models in this work. Using the LSTM and BiLSTM techniques, a sequential model is generated. In this implementation, a convolutional one-dimensional layer with 32 filters, a 3x3 kernel size, a padding that preserves the same dimension on both the left and right sides of the data, and an activation function denoted `relu' are utilized. The core layers were then built using bidirectional LSTM for CNNBiLSTM and LSTM for CNN-LSTM with 128 and 64 neurons, respectively. An output dense layer with a single neuron and sigmoid activation function is also added. These models are utilized to precisely train each label as a classifier chain. In light of this, CNN-BiLSTM has demonstrated superior performance across all rating categories.

\begin{table}[h]
\begin{center}
\begin{minipage}{130pt}
\caption{Overall accuracy score of DL algorithms}\label{tab7}
\begin{tabular}{lr}
\toprule
Model  & Accuracy (\%) \\
\midrule
LSTM            & 78.44                  \\
\textbf{BiLSTM} & \textbf{79.14}         \\
CNN-BiLSTM      & 78.70                  \\
CNN-LSTM        & 78.29                 \\
\botrule
\end{tabular}
\footnotetext{Bold value specifies the highest result among DL algorithms.}
\end{minipage}
\end{center}
\end{table}

\begin{figure}[h]%
\centering
\includegraphics[width=0.9\textwidth]{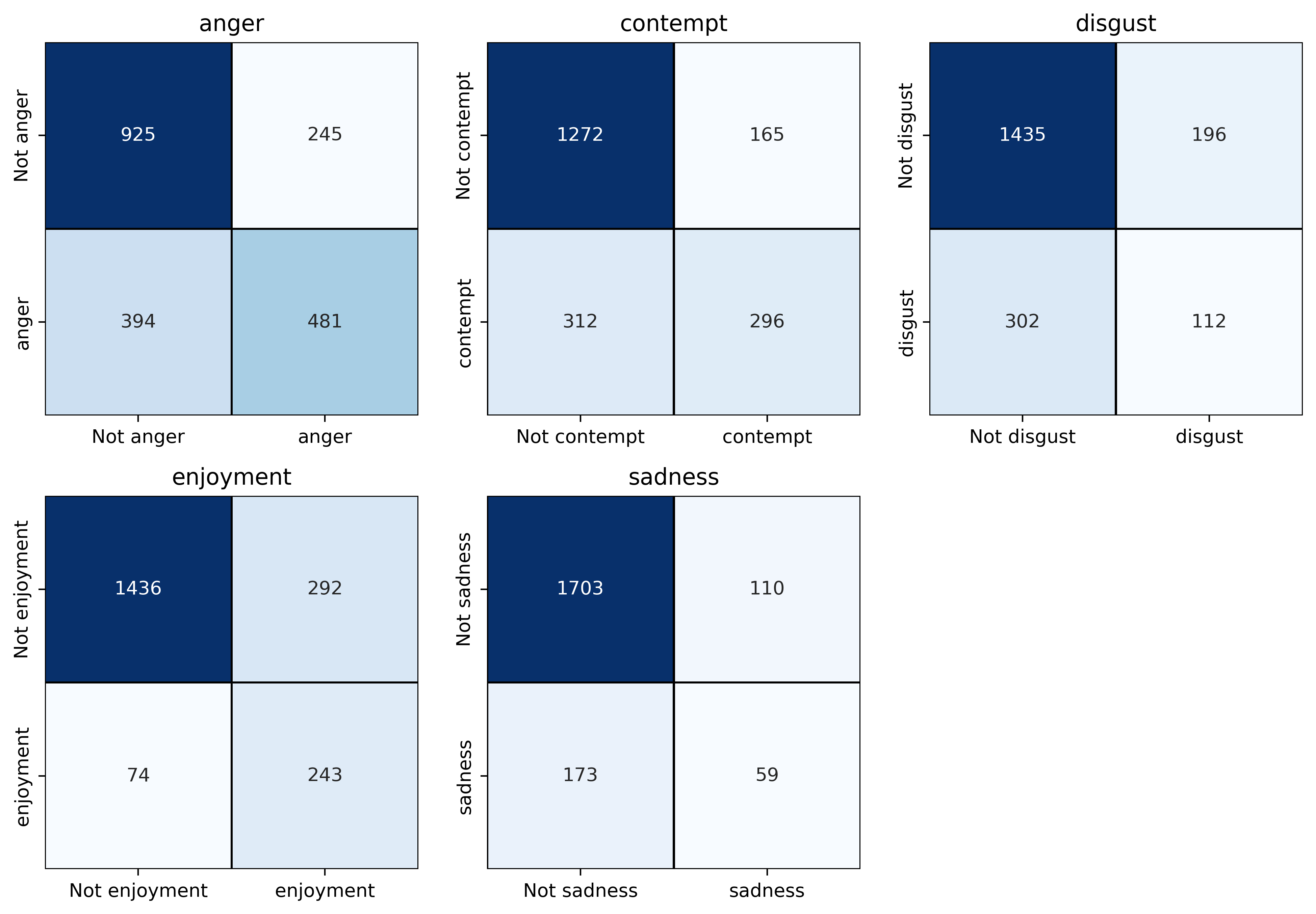}
\caption{Confusion matrix of BiLSTM that is best performer among DL approaches. Each box indicates the number of correctly and incorrectly identified emotions.}\label{fig5}
\end{figure}

Finally, in terms of DL methods for multi-label ER, BiLSTM fared the best. The confusion matrix generated by implementing the BiLSTM approach is depicted in \autoref{fig5}, and the overall accuracy of the DL methods is shown in \autoref{tab7}. It reveals that the algorithm can predict emotions adequately; but, its forecasts for rage, scorn, and disgust are wrong. Due to this shortcoming, the accuracy of this method is inferior to that of MNB.

\subsubsection{Evaluation of BERT approaches}\label{subsubsec3.3.3}
BERT is a transformer-based model. It was created in 2018 by Google AI Language researchers as a one-stop solution for most common language tasks, including sentiment classification and named entity classification \cite{Das2021}. BERT employs an attention pathway that establishes contextual associations between words (or sub-words) in a text. BanglaBERT is the first Bangla Language Understanding Benchmark (BLUB) developed in Bangladesh \cite{Islam2020, Das2021}. However, frequently used pre-trained BERT approaches such as BanglaBERT, multilingual BERT (mBERT), Bangla-Bert-Base, and Bangla-Electra are used in this study to understand the multi-label ER. The outcomes based on different emotions are illustrated in \autoref{tab5}.

\begin{sidewaystable}
\sidewaystablefn%
\begin{center}
\begin{minipage}{\textheight}
\caption{Precision, recall, and F1 score of BERT approaches}
\label{tab5}
\setlength{\tabcolsep}{3pt} 
\renewcommand{\arraystretch}{1} 
\begin{tabular*}{\textheight}{lcccc|cccc|cccc}
\toprule%
& \multicolumn{4}{c}{Precision (\%)}& \multicolumn{4}{c}{Recall (\%)} & \multicolumn{4}{c}{F1 Score} \\\cmidrule{2-5}\cmidrule{6-9}\cmidrule{10-13}%
Emotions & \begin{tabular}[c]{@{}c@{}}Bangla \\Electra\end{tabular} & \begin{tabular}[c]{@{}c@{}}Bangla-\\Bert-Base\end{tabular} & mBERT & \begin{tabular}[c]{@{}c@{}}Bangla\\BERT\end{tabular} & \begin{tabular}[c]{@{}c@{}}Bangla \\Electra\end{tabular} & \begin{tabular}[c]{@{}c@{}}Bangla-\\Bert-Base\end{tabular} & mBERT & \begin{tabular}[c]{@{}c@{}}Bangla\\BERT\end{tabular} & \begin{tabular}[c]{@{}c@{}}Bangla \\Electra\end{tabular} & \begin{tabular}[c]{@{}c@{}}Bangla-\\Bert-Base\end{tabular} & mBERT & \begin{tabular}[c]{@{}c@{}}Bangla\\BERT\end{tabular}\\
\midrule
\textbf{Anger}                     & \textbf{71.92}          & \textbf{72.98}                                                       & 70.7                                                                  & 69.56               & 70.74                   & \textbf{72.82}            & \textbf{67.69}             & 72.6                & 0.71                    & \textbf{0.73}             & \textbf{0.69}              & 0.71                \\
\textbf{Contempt}                  & 54.37                   & \textbf{66.85}                                                       & 61.1                                                                  & 60.4                & 52.59                   & \textbf{61.21}            & \textbf{53.62}             & 57.07               & 0.53                    & \textbf{0.64}             & \textbf{0.57}              & 0.59                \\
\textbf{Disgust}                   & \textbf{0.0}            & \textbf{45.69}                                                       & 37.57                                                                 & 44.33               & \textbf{0.0}            & \textbf{41.67}            & 15.48                      & 10.24               & \textbf{0.0}              & \textbf{0.44}             & 0.22                       & 0.17                \\
\textbf{Enjoyment}                 & 83.79                   & 84.33                                                                & \textbf{72.03}                                                        & \textbf{89.13}      & 69.74                   & \textbf{74.34}            & 55.92                      & \textbf{67.43}      & \textbf{0.76}           & \textbf{0.79}             & 0.63                       & 0.77                \\
\textbf{Sadness}                   & 0.0                     & \textbf{52.88}                                                       & 42.03                                                                 & 0.0                 & \textbf{0.0}            & \textbf{41.91}            & 12.03                      & 0.0                 & \textbf{0.0}            & \textbf{0.47}             & 0.19                       & 0.0                \\
\botrule
\end{tabular*}
\footnotetext{Bold values indicate the highest precision, recall, and F1 score among the algorithms for the classified emotions.}
\end{minipage}
\end{center}
\end{sidewaystable}

From the basic transformers library, BERT approaches are built using a method called as multi-label classification model. The pre-trained models are used with certain parameters, including reprocessing of input data even if cached data exists, cached features with a false boolean value during evaluation, and a value of 3 for the number of epochs. The data is then used to train these models for assessment. In the assessment step, the predictions provided by the models for each emotion label are assessed in terms of precision, recall, f1 score, and total accuracy. Consequently, it demonstrates the application of classifier chain approaches in multi-label classification. In addition, the overall performance of all BERT algorithms is comparable. In terms of accuracy, recall, and f1 score, Bangla-Bert-Base has the best overall performance superiority. Even though BanglaBert obtained the greatest accuracy score for pleasure, BanglaBert-Base received the highest f1 score for that feeling. Furthermore, Bangla-Bert-Base beat every other algorithm utilized in this investigation. It achieves 83.23\% accuracy, which is comparable to the ML algorithm MNB. The confusion matrix in \autoref{fig5} demonstrates that Bangla-BertBase is more accurate than all other algorithms, despite its accuracy being comparable to that of the ML method.Therefore, the total precision of all BERT methods is illustrated in \autoref{tab8}

\begin{table}[h]
\begin{center}
\begin{minipage}{150pt}
\caption{Overall accuracy score of BERT approaches}\label{tab8}
\begin{tabular}{lr}
\toprule
Model          & Accuracy (\%) \\
\midrule
Bangla-Electra            & 81.79                  \\
\textbf{Bangla-Bert-Base} & \textbf{83.23}         \\
Multilingual Bert         & 80.92                  \\
BanglaBert                & 82.21                 \\
\botrule
\end{tabular}
\footnotetext{Bold value specifies the highest result among BERT approaches.}
\end{minipage}
\end{center}
\end{table}

\begin{figure}[h]%
\centering
\includegraphics[width=0.9\textwidth]{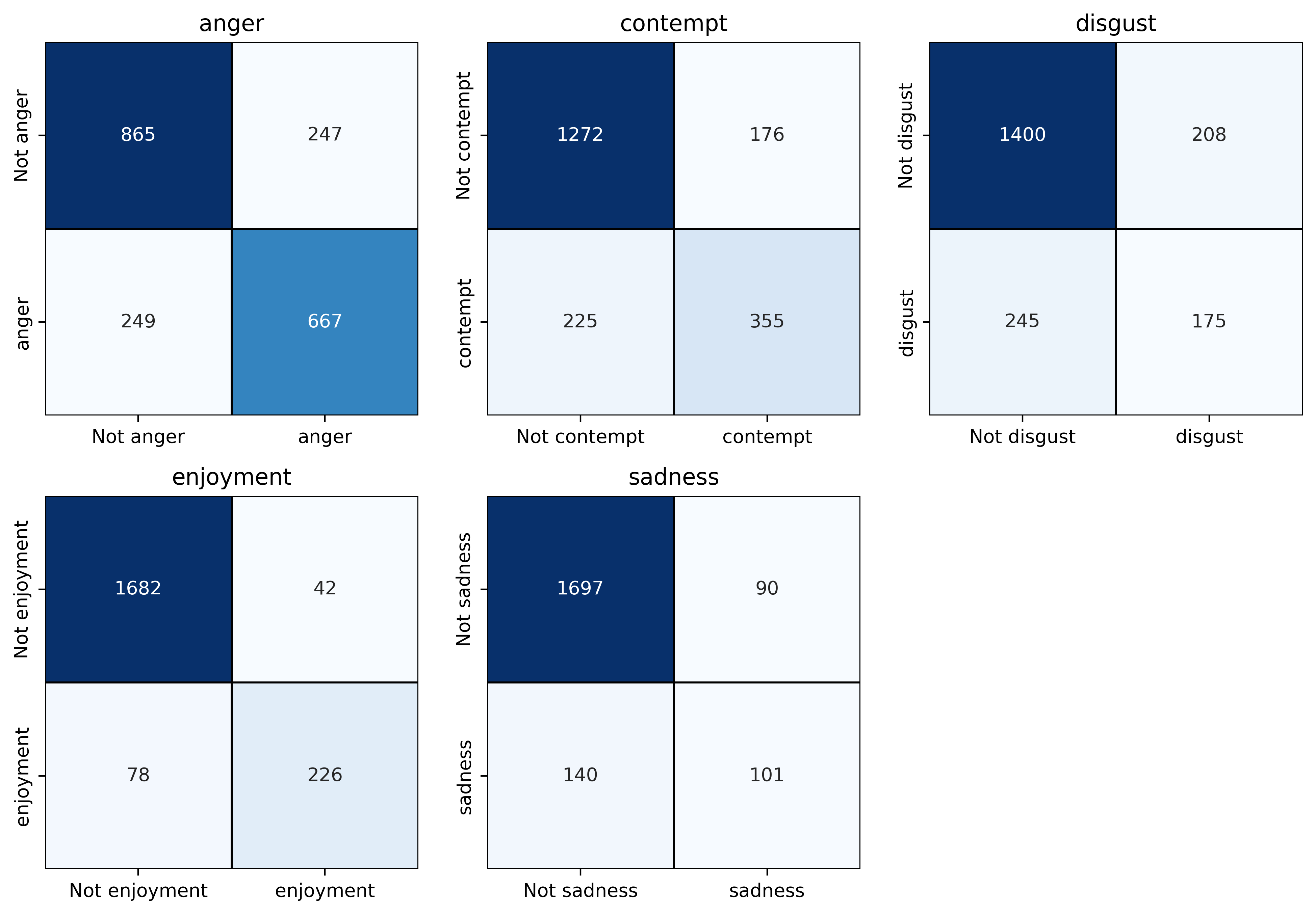}
\caption{Confusion matrix of Bangla-Bert-Base that is best performer among BERT approaches. Each box indicates the number of correctly and incorrectly identified emotions.}\label{fig6}
\end{figure}

\begin{figure}[h]%
\centering
\includegraphics[width=0.9\textwidth]{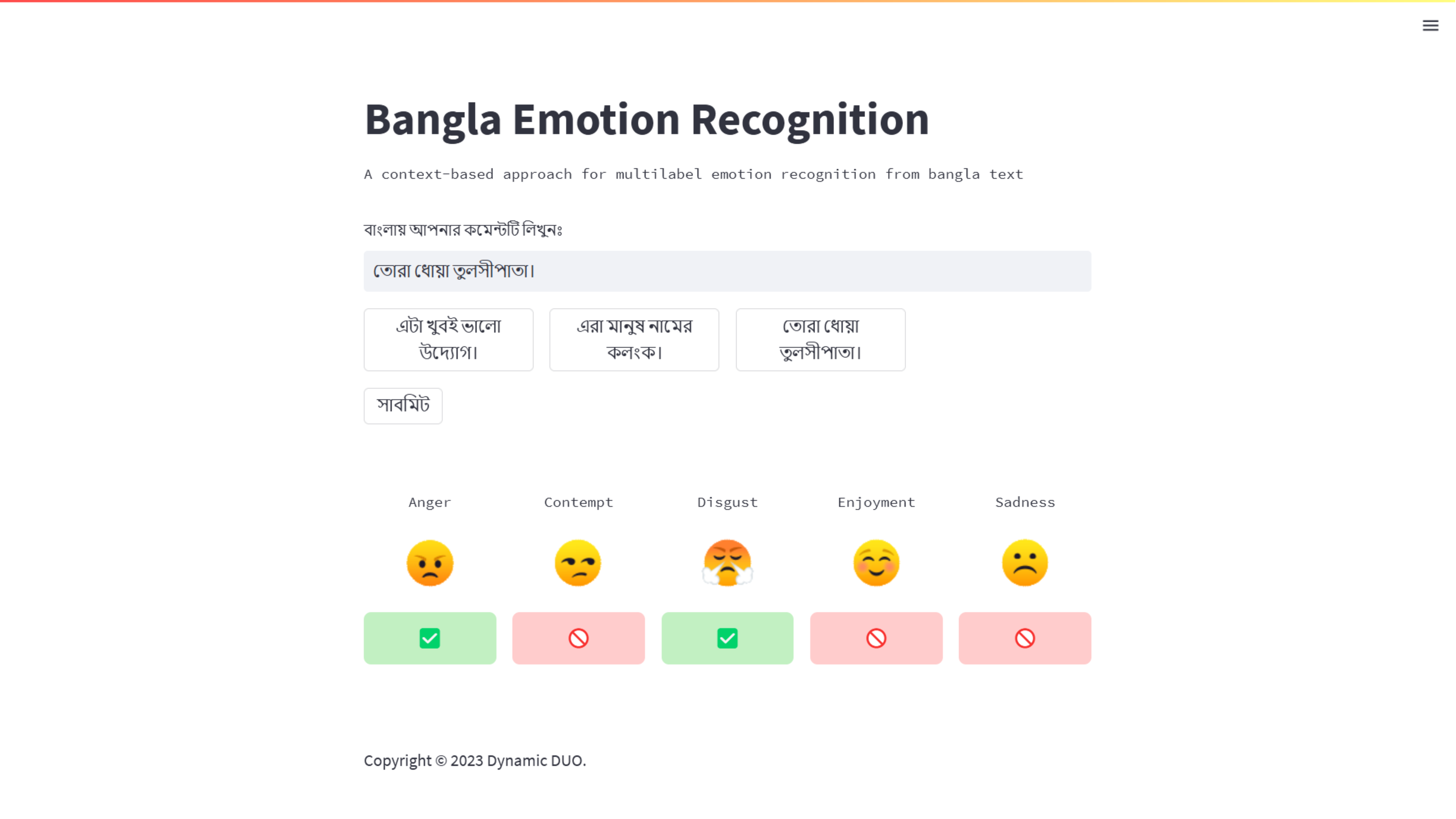}
\caption{Interface of web application where some suggested captions are shown and one of them is clicked for detecting emotion. The pre trained model shows the identified emotions of the selected comment.
}\label{fig7}
\end{figure}

Using the highest performing model described in this paper, a web service for identifying emotions in Bangla text has finally been developed. The implementation and source code of the application may be accessed on GitHub at \footnote{\url{https://github.com/Shammi179/Multilabel_Bangla_ER_webapp}}. The web application is constructed using the most performing version of Bangla-best-base. The user is required to enter a Bangla remark in the input box, following which the pre-trained Bangla-bert-base model will assess the sentiment of the supplied statement. The UI of the web application is exemplified in \autoref{fig7}.

In this exemplary illustration, a user has entered a comment, and the pre-trained model has predicted three related emotions. For the development of the Streamlit application, a framework is applied, and a trained model is used to reliably detect emotions. In addition, the interface includes three sample instances from which the user may rapidly pick one to observe the result. Alternatively, a user may input a Bangla sentence, and the pre-trained model will identify the emotions based on the training. Therefore, the application's UI is more straightforward, and animated emojis have been integrated to enhance the application's aesthetic and comprehension.

\section{Discussion}\label{sec4}
Given that Bangla is a resource-constrained language, this study aims to overcome the limitations of this domain and present a unique strategy for multi-label ER. To address the issue of insufficient annotated corpora, a comprehensive corpus called `MONOVAB' has been compiled in this study. This article proposes an automated way of acquiring Facebook data using a data scraper. Furthermore, it focuses on building a methodological framework for multi-label ER in Bangla utilizing the annotated and compiled dataset. The obtained data has been processed using conventional preprocessing and feature extraction techniques. Following this, ML, DL, and transformer-based algorithms are investigated, as they are the most used methodologies for emotion recognition [4]. MNB has performed the best among the ML algorithms LR, MNB, SVM, RF, and KNN, with an accuracy of 82.68\%. There are several reasons for this algorithm's superior performance, apart from its easier implementation. Consequently, the classification of textual data and real-time applications can benefit from using this. Therefore, LSTM, BiLSTM, CNN-LSTM, and CNN-BiLSTM are implemented from the DL domain, where BiLSTM outperformed other DL algorithms with an accuracy of 79.14\%. Bidirectional LSTM has been added to traditional LSTM to increase the performance of sequence classification models. Bidirectional LSTMs are applied to train two LSTMs rather than one on the input sequence. The first operation is performed on the regular input sequence, while the second operation is performed on the input sequence's mirror image. This has supplied the network with extra context, leading to a faster increase in its understanding of the issue.

BERT algorithms have been incorporated because this technique has gained prominence over the past few years. The Bangla-Electra, Bangla-Bert-Base, Multilingual Bert, and Bangla Bert algorithms are analyzed based on BERT techniques. Compared to other algorithms, Bangla-Bert-Base has the highest accuracy at 83.23 percent, which is remarkable. BERT is able to manage more text than other systems. It employs a basic strategy based on previously learned models. It has the ability to tailor info to the individual environment and circumstance encountered. Therefore, the model's output is sufficient and may be used for further multi-label ER in Bangla. Finally, a web application is developed to illustrate the performance of the pre-trained Bangla-Bert-Base model. The application is developed using the Streamlit framework, and a pre-trained model is utilized to accurately identify the emotions. Three sample instances are also included in the UI (User Interface) of this online application, which may be used to identify emotions, making it simple for users to select one and see the results. Additionally, a user can type some texts in Bangla, and the pre-trained model will recognize the emotions in accordance with the text. The application's UI is simpler to understand, and animated emojis have been added for better look and comprehension. 

\section{Conclusions}\label{sec7}

In terms of linguistic resources, Bangla is regarded as a low-resource language due to the absence of adequate tools as well as correctly classified data. This work focuses towards minimizing the shortcomings with the formation of an annotated corpus from the data obtained from the newspaper’s Facebook pages. In this convergence, an automated data scrapper has been developed to ease the further processing of Bangla text. The efficiency of the data scrapper can be visualized by the set collected in this work having a total of 1,36,583 comments. The annotation phase included the use of context-based annotation, which aided in identifying the intended tone of the text. Based on these context-based annotations, a cleaned, compiled dataset has been built namely `MONOVAB'. Then, the context-based data is used to complete multi-label ER, an almost untouched approach for Bangla Text. Multiple approaches have been used to determine the connotations, including ML, DL, and BERT, but Bangla-Bert-Base has shown to be the most prominent, with an accuracy of 83.23\%. Not only in terms of accuracy but also the algorithm's other metrics, recall, and F1 score, are likewise satisfactory. The whole set of collected data can’t be adapted in the annotated corpora due to the lack of annotations which can be further utilized to develop a lexicon-based approach.



\section{Declaration}
The preprint version is available at arxiv\cite{banshal2023monovab}. 
\bibliography{sn-bibliography}


\end{document}